\documentclass[journal]{IEEEtran}

\usepackage{amsmath,amsfonts}
\usepackage{algorithmic}
\usepackage{algorithm}
\usepackage{array}
\usepackage[caption=false,font=normalsize,labelfont=sf,textfont=sf]{subfig}
\usepackage{textcomp}
\usepackage{stfloats}
\usepackage{amsmath}

\usepackage{url}
\usepackage{verbatim}
\usepackage{graphicx}
\usepackage{cite}
\usepackage{booktabs}
\usepackage{pgfplots}
\usetikzlibrary{patterns}
\usepackage{multirow}
\usepackage{makecell}
\newcolumntype{d}[1]{D{.}{.}{#1}}

\usepackage[normalem]{ulem}

\newcommand{\ORplayer}{\pi_{\vee}}
\newcommand{\ANDplayer}{\pi_{\wedge}}
\pgfplotsset{compat=1.18}

\newif\ifauhor\auhorfalse

\begin{document}

\title{A Study of Solving Life-and-Death Problems in Go Using Relevance-Zone Based Solvers
}

\author{Chung-Chin Shih, \IEEEmembership{Member, IEEE}, Ti-Rong Wu, \IEEEmembership{Senior Member, IEEE}, Ting Han Wei, Yu-Shan Hsu, Hung Guei, \IEEEmembership{Member, IEEE}, and I-Chen Wu, \IEEEmembership{Senior Member, IEEE}%
\thanks{This research is partially supported by the National Science and Technology Council (NSTC) of the Republic of China (Taiwan) under grant numbers 113-2221-E-001-009-MY3, 113-2634-F-A49-004, 114-2221-E-A49-005, 114-2221-E-A49-006, 114-2628-E-A49-002, and 114-2634-F-A49-004. (Corresponding author: Ti-Rong Wu)}

\thanks{Chung-Chin Shih and Ti-Rong Wu are with the Institute of Information Science, Academia Sinica, Taipei, Taiwan (e-mail: rockmanray@iis.sinica.edu.tw; tirongwu@iis.sinica.edu.tw).}
\thanks{Ting Han Wei is with the School of Informatics, Kochi University of Technology, Kami City, Japan (e-mail: tinghan.wei@kochi-tech.ac.jp).}
\thanks{Yu-Shan Hsu is with the Graduate Degree Program of Robotics, National Yang Ming Chiao Tung University, Hsinchu, Taiwan (e-mail: samox0918@gmail.com).}
\thanks{Hung Guei is with the Institute of Information Science, Academia Sinica, Taipei, Taiwan (e-mail: hguei@iis.sinica.edu.tw).}
\thanks{I-Chen Wu is with the Department of Computer Science, National Yang Ming Chiao Tung University, Hsinchu, Taiwan (e-mail: icwu@cs.nycu.edu.tw).}
}

\markboth{A Study of Solving Life-and-Death Problems in Go Using Relevance-Zone Based Solvers}
{Shell \MakeLowercase{\textit{et al.}}: A Sample Article Using IEEEtran.cls for IEEE Journals}

\maketitle

\begin{abstract}
This paper analyzes the behavior of solving Life-and-Death (L\&D) problems in the game of Go using current state-of-the-art computer Go solvers with two techniques: the Relevance-Zone Based Search (RZS) and the relevance-zone pattern table.
We examined the solutions derived by relevance-zone based solvers on seven L\&D problems from the renowned book "\textit{Life and Death Dictionary}" written by Cho Chikun, a Go grandmaster, and found several interesting results.
First, for each problem, the solvers identify a relevance-zone that highlights the critical areas for solving.
Second, the solvers discover a series of patterns, including some that are rare.
Finally, the solvers even find different answers compared to the given solutions for two problems.
We also identified two issues with the solver: (a) it misjudges values of rare patterns, and (b) it tends to prioritize living directly rather than maximizing territory, which differs from the behavior of human Go players.
We suggest possible approaches to address these issues in future work.
Our code and data are available at https://rlg.iis.sinica.edu.tw/papers/study-LD-RZ.
\end{abstract}


\section{Introduction}
\label{sec:intro}
AlphaZero \cite{silver_mastering_2017} and MuZero \cite{schrittwieser_mastering_2020} have made it possible to train superhuman agents without human knowledge. 
However, there are still imperfections in agents trained with these algorithms.
For example, ELF OpenGo often mishandled ladders, a common Go pattern that is taught to novice human players \cite{tian_elf_2019}. 
In another example, several research papers have pointed out ways to exploit KataGo \cite{wu_accelerating_2020a}, a current state-of-the-art open source computer Go program, by applying adversarial policies \cite{lan_are_2022,wang_adversarial_2023}.
In addition, attempts to analyze notoriously difficult problems (such as one included in the famous book \textit{Igo Hatsuyoron}) with AlphaZero-like programs have not yielded definitive results \cite{wu_deeplearning_2019}.

\begin{figure}[t]
\centering
\subfloat[]{\includegraphics[height=1.55cm]{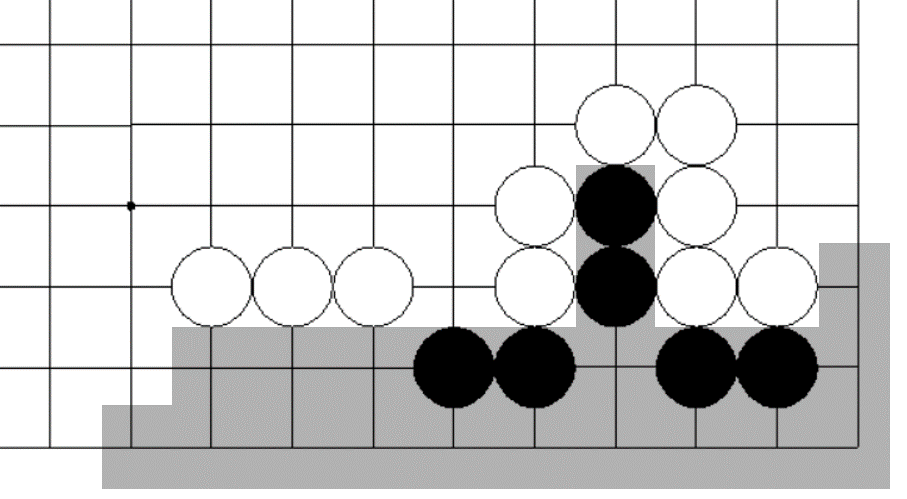}\label{fig:intro_a}}
\subfloat[]{\includegraphics[height=1.55cm]{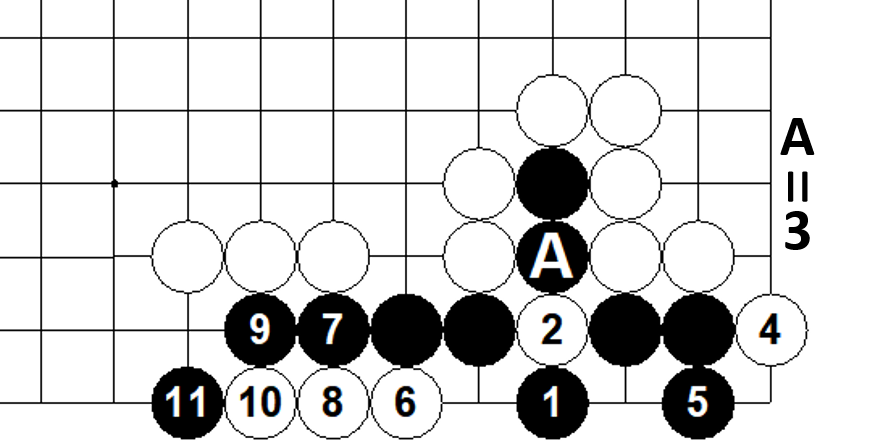}\label{fig:intro_b}}
\caption{
The problem is Black to play and live. (a) The minimum restricted area for both players to verify the solutions. (b) A position that might be encountered in the solutions.
}
\label{fig:intro}
\end{figure}

Ground truths can provide an exact basis for comparison when evaluating and interpreting AlphaZero-like programs.
Proofs in the form of solution trees \cite{stockman_minimax_1979, pijls_game_2001}, such as solutions to chess endgames or \textit{Life-and-Death (L\&D)} problems in Go\footnote{In Go, two players alternately place black and white stones on the board intersections, and a group of connected stones is captured when all its adjacent intersections, called liberties, are all occupied by the opponent.}, are clear ways of presenting such an optimal strategy.
In previous research, Kishimoto and M{\"u}ller \cite{kishimoto_search_2005} introduced an efficient method to solve L\&D problems under limited time or node constraints.
However, this method relies on human knowledge to restrict the search space. 
For example, in Fig. \ref{fig:intro}, an L\&D problem, where Black plays first and is expected to live.
Go players understand that moves outside of the shaded area do not contribute to Black's survival.
It is therefore critical to reduce the shaded area as much as possible to speed up the search, but not so much that the result may be incorrect, e.g., to the point where the shaded area excludes any of the played moves in Fig. \ref{fig:intro_b}.
Designating this reduced search space requires Go knowledge and human intervention.

Shih et al. \cite{shih_novel_2022} proposed two methods to improve L\&D analysis efficiency. 
First, \textit{Relevance-Zone Based Search (RZS)} automatically identifies this limited search space during the search process, therefore removing the need for manual intervention. 
Moves played by the opponent outside the RZ do not affect the winning player's strategy, and can therefore be omitted without affecting the outcome. 
Next, to address the tendency that heuristic agents prioritize win rate over depth of play, they introduced the \textit{Faster-to-Life (FTL)} neural network.
This incentivizes the winning player to win using fewer moves, which also reduces the depth of the search. 
These two methods were applied to, among other benchmarks, 106 $19 \times 19$ L\&D problems.
The experiments indicated significant improvements over the previous method proposed by Kishimoto and M{\"u}ller \cite{kishimoto_search_2005} -- from 11 solved problems to 68.
This result was further improved with the development of a radix tree-based RZ transposition table \cite{shih_localpattern_2023}, further increasing the number of solved problems from 68 to 83.

This paper focuses on analyzing the 83 L\&D problems solved in this set of $19 \times 19$ Go problems and discussing the problems encountered by the solvers during the search.

The remaining sections of this paper are organized as follows. Section \ref{sec:background} introduces the background of the techniques discussed in this paper.
Section \ref{sec:solverandproblems} discusses the solvers used. 
Section \ref{sec:analysis} selects representative problems from the solved L\&D problems for analysis and discussion. 
Finally, Section \ref{sec:discussion} provides the discussion and concluding remarks.

\begin{figure}[t]
\centering
\subfloat[$p_a$]{\includegraphics[width=0.245\columnwidth]{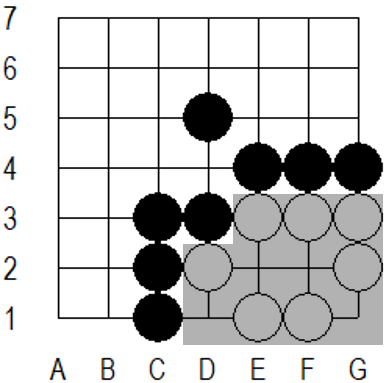}\label{fig:ex_a}}
\subfloat[$p_{b'}$]{\includegraphics[width=0.245\columnwidth]{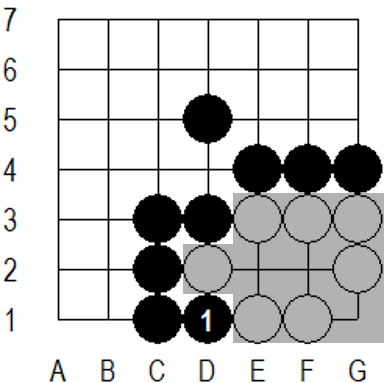}\label{fig:ex_b}}
\subfloat[$p_{c'}$]{\includegraphics[width=0.245\columnwidth]{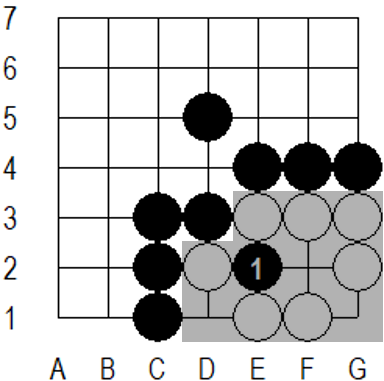}\label{fig:ex_c}}
\subfloat[$p_{d'}$]{\includegraphics[width=0.245\columnwidth]{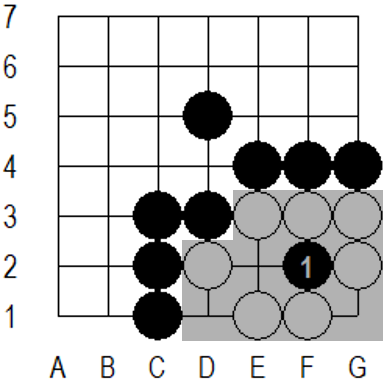}\label{fig:ex_d}}
\\
\subfloat[$p_*$]{\includegraphics[width=0.245\columnwidth]{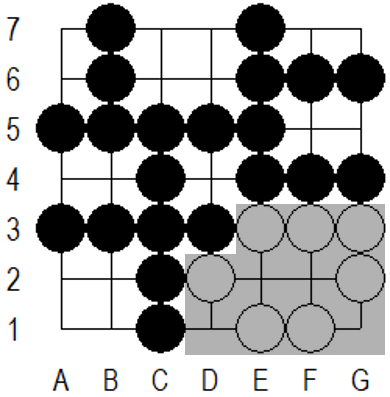}\label{fig:ex_e}}
\subfloat[$p_b$]{\includegraphics[width=0.245\columnwidth]{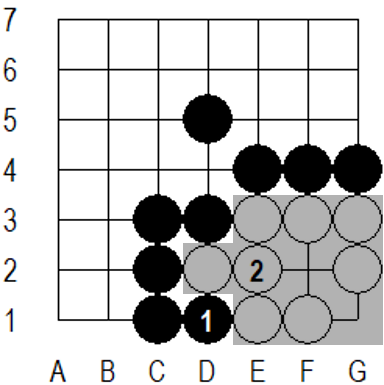}\label{fig:ex_f}}
\subfloat[$p_c$]{\includegraphics[width=0.245\columnwidth]{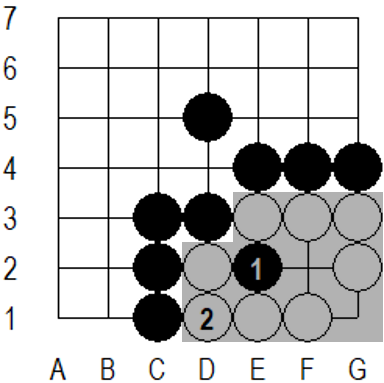}\label{fig:ex_g}}
\subfloat[$p_d$]{\includegraphics[width=0.245\columnwidth]{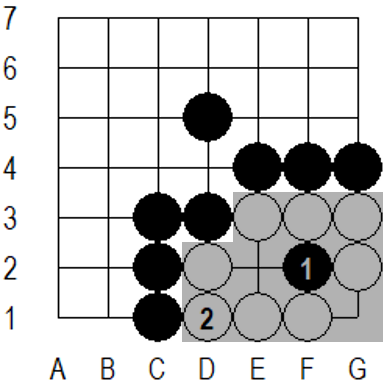}\label{fig:ex_h}}
\caption{
An example to illustrate relevance-zones in Go, modified from \cite{shih_novel_2022}.
}
\label{fig:examples}
\end{figure}

\section{Background}
\label{sec:background}

\subsection{Life-and-Death Problems}
\label{subsec:rules_LAD}

L\&D problems are sometimes called \textit{tsumego}. 
They can be viewed as a subgame of Go. 
Although Go is a game with symmetric rules in which both players share the same objective, we present here an abstract description that focuses on the perspective of one player. 
In a two-player game where one player aims to accomplish a specific goal while the other player aims to prevent it, the player seeking the goal is called the OR-player ($\ORplayer$), and the player trying to prevent the goal is known as the AND-player ($\ANDplayer$).
Winning, from the perspective of the OR-player, is equivalent to successfully achieving the goal.
Given a position and some \textit{crucial} stones \cite{kishimoto_search_2005} of $\ORplayer$, $\ORplayer$ must guarantee safety for any of these crucial stones while $\ANDplayer$ tries to capture all crucial stones. 
Another simple example of an L\&D problem is shown in Fig. \ref{fig:ex_a}, where it is Black’s ($\ANDplayer$) turn to play and White ($\ORplayer$) can live. 
White's goal is ensuring safety of any of the crucial stones (all three blocks in the shaded area); Black's goal is capturing all these white blocks.

A set of blocks is said to be unconditionally alive (UCA) \cite{benson_life_1976} if the opponent cannot capture it even if the opponent can play unlimited consecutive stones on the board. 
In Fig. \ref{fig:ex_f}, \ref{fig:ex_g}, and \ref{fig:ex_h}, the white blocks have achieved UCA and are thus terminal positions in the search to solve this problem.

\subsection{Relevance-Zone Based Search}
\label{subsec:RZS}

A zone $z$ refers to a region of intersections on a board, as illustrated by the shaded area in Fig. \ref{fig:ex_a}.
For a position $p$, there exists a set of similar positions $p_*$ where the stones inside $z$ are the same as those in $p$, as illustrated in Fig. \ref{fig:ex_a} and Fig. \ref{fig:ex_e}. 
Such a zone $z$ can be called a relevance-zone (RZ) with respect to a winning position $p$ (for $\ORplayer$), if all $p_*$ are also winning for $\ORplayer$.
In other words, an RZ contains all subsequent moves that are relevant to winning. 
For Go, there will always exist at least one RZ with respect to a winning position. 
In the most extreme case, the RZ is the entire board (i.e., no search reduction will be attempted).

Relevance-Zone Based Search (RZS) \cite{shih_novel_2022} is a search algorithm that takes advantage of RZs, which is used to prune irrelevant move in the search space.
In Fig. \ref{fig:examples}.
For position $p_a$, if Black plays at 1 as in $p_{b'}$, $p_{c'}$, and $p_{d'}$, White can reply 2 to achieve UCA and win as in $p_b$, $p_c$, and $p_d$, respectively. 
Since positions $p_b$, $p_c$, and $p_d$ are UCA for white stones, we can designate the RZs as the shaded intersections, which we denote by $z_b$, $z_c$, and $z_d$, respectively. 
For their preceding positions $p_{b'}$, $p_{c'}$, and $p_{d'}$, White wins by playing at 2, so therefore their corresponding RZs are also $z_b$, $z_c$, and $z_d$, respectively. 
The stone configurations inside these zones form \textit{RZ patterns}. 
Since Black 1 in Fig. \ref{fig:ex_f} is outside of $z_b$, it can be considered irrelevant (sometimes referred to as a null move). 
White can win by replying at E2 for all Black moves outside $z_b$. 
The zone $z_a$ is the union of $z_b$, $z_c$, and $z_d$, as shaded in Fig. \ref{fig:ex_a}.
Detailed illustrations and the proof of correctness for replaying winning strategies in an RZ are omitted in this paper. 
RZS can therefore automatically reduce the search space, significantly improving L\&D problem analysis efficiency \cite{shih_novel_2022}.

\subsection{Faster-to-Life Network}
\label{subsec:ftl}

Faster to Life (FTL) \cite{shih_novel_2022} is an AlphaZero-like training method for a deep neural net heuristic that prefers moves that lead to faster wins instead of higher win rates.
The motivating use case for FTL is a variant of Go called Killall Go, where the victory condition was that White simply had to achieve UCA anywhere on a reduced board.
In original AlphaZero training, territory is counted at the end of a game, and the win/loss is awarded to each player accordingly.
In FTL, White ($\ORplayer$) wins if it can achieve UCA for any number of its stones within $d$ moves from the current position.
For example, with $d = 20$, White wins only if it is able to reach UCA within 20 moves, i.e. even if White achieves UCA in 21 moves, it counts as a loss.
Given multiple values of $d$, the problem setting is similar to determining win/loss with multiple komi values, which can be handled with multi-labelled value networks \cite{wu_multilabeled_2018}.

Given a position $p$, the value network head of FTL outputs a set of additional $d$-win rates (the rates of winning within $d$ moves, ranging from -1 to 1), where $d$ ranges from 1 to a sufficiently large number. 
Namely, for all positions $p$ during the self-play portion of AlphaZero training, if White lives by UCA at the $d$-th move from the current position, one more $d$-win was counted for all $d'\geq d$, but not for all $d'< d$.  
These $d'$-wins are used to update the set of $d$-win rates of $p$. 
This incentivizes White to win faster if their $d$-win rates are high for low values of $d$.
Another similar method is the Proof-Cost Network \cite{wu_alphazerobased_2021}, which aims to solve problems with smaller solution trees.

\subsection{Relevance-Zone Pattern Table}
\label{subsec:rzt}

Following the definition of RZs, positions with different stone configurations that match the same RZ patterns within $z$ are also wins for $\ORplayer$. 
In Fig. \ref{fig:ex_a}, the RZ pattern in $z_a$ can be reused to indicate that the position $p_*$ in Fig. \ref{fig:ex_e} is also a win. 

To facilitate future lookups and find matching patterns for a given position, a table can be implemented to store various winning patterns within $z$. 
Traditional lookup schemes such as transposition tables using Zobrist hashing \cite{zobrist_new_1990} are a specific case of this pattern matching mechanism, where $z$ is the entire board. 
Shih et al. proposed a pattern table based on radix trees named the RZ Pattern Table \cite{shih_localpattern_2023}. 
This table collects patterns that are generated during RZS and provides efficient lookup schemes. 
With this table, the performance of RZS was further improved \cite{shih_localpattern_2023}. 

\section{The Solvers and Problem Sets}
\label{sec:solverandproblems}

This section introduces the L\&D solver and problem set we investigate in this paper. 

The solver is based on the CGI \cite{wu_accelerating_2020} program. 
The FTL networks were trained using 5 residual blocks and 64 filters, where a total of 2,500,000 self-play games were generated during training, with 400 simulations for each move. 
The learning rate was initially set to 0.02 for the first 1,500,000 self-play games and then reduced to 0.002 for finetuning in later training. 
The network was trained using a total of 2,946 problems from a L\&D book titled “The Training of Life and Death Problems in Go” \cite{shao_training_1991}. 
During training, self-play games end once one of the crucial stones reaches UCA, or when all crucial stones are captured. 

We analyze the solving process with two different solvers. \textit{RZS-TT} uses a traditional transposition table \cite{marsland_review_1986}; \textit{RZS-PT} uses the RZ pattern table \cite{shih_localpattern_2023} described in subsection \ref{subsec:rzt}.
Although \cite{shih_localpattern_2023} shows that RZS-PT performs better than RZS-TT, comparing the two provides deeper insight into how pattern-based reuse affects search behavior.
 
We evaluated the two solvers on a collection of 106 L\&D problems, selected from the renowned book "\textit{Life and Death Dictionary}" written by Cho Chikun \cite{cho_life_1987}. 
Among the 106 problems, 68 can be solved by RZS-TT in five minutes, while RZS-PT was able to solve 83 problems within the same time limit. 
The experiments were conducted on a single NVIDIA GTX 1080Ti GPU and an Intel(R) Xeon(R) CPU E5-2683 v3, during which approximately 200,000 nodes were typically searched within five minutes.
To compare the performance of the two solvers, both solvers needed to solve the same 83 problems. 
RZS-TT was allowed to continue searching without time limits for the remaining 15 problems.
The results show that RZS-PT was able to solve problems 4.74 times faster compared to RZS-TT.
In summary, both RZS-TT and RZS-PT are suitable for L\&D analysis, with RZS-PT consistently outperforming RZS-TT with significant improvements in terms of simulation counts.

\section{Problem Analysis}
\label{sec:analysis}

This section investigates the solving behavior of RZS-TT and RZS-PT.
In this analysis, seven problems were selected by an amateur 6-dan Go player from the 83 $19 \times 19$ problems solved by both solvers. 
The selection criteria aimed to include cases that highlight different aspects of solver behavior, namely: (a) problems where the search region extends farther from the original stones in the book (Problems 1 and 2), (b) problems in which the solver's answers differ from the book solutions (Problems 3 and 4), and (c) problems featuring interesting or uncommon patterns (Problems 5, 6, and 7).
Two problems are from Volume 1, and five problems are from Volume 2 of the book \cite{cho_life_1987}. 
We discussed the issues encountered while solving these problems, and we also discovered interesting results.
 
In subfigure (a) of each problem, an A or both A and B are marked to indicate the answers that the solvers found or the book provided. 
In the last subfigure of each problem, the shaded area represents the RZ of the problem by the solvers.

\subsection{Problem 1}
\label{subsec:problem1}

\begin{figure}[t]
\centering
\subfloat[]{\includegraphics[width=0.33\columnwidth]{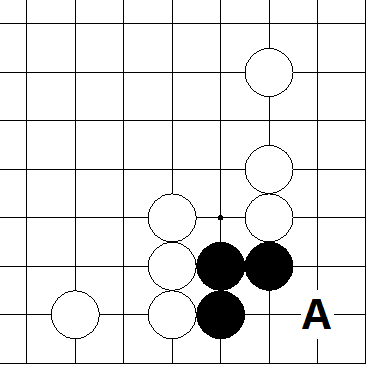}\label{fig:prob1_a}}
\subfloat[]{\includegraphics[width=0.33\columnwidth]{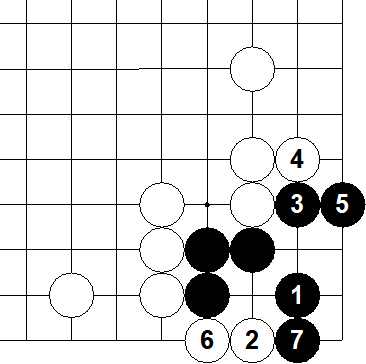}\label{fig:prob1_b}}
\\
\subfloat[]{\includegraphics[width=0.33\columnwidth]{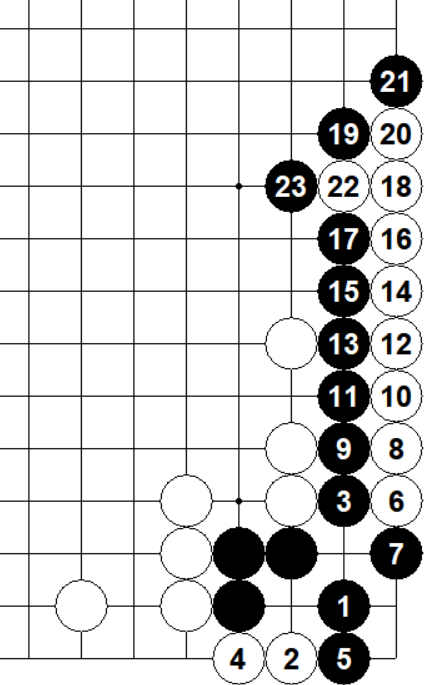}\label{fig:prob1_c}}
\subfloat[]{\includegraphics[width=0.33\columnwidth]{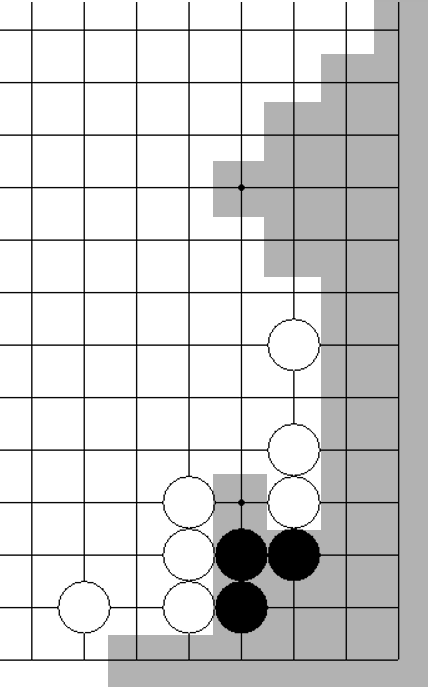}\label{fig:prob1_d}}
\caption{
Problem 1, from Volume 1, page 88 of the book.
}
\label{fig:problem1}
\end{figure}

This problem shown in Fig. \ref{fig:prob1_a} is from page 88 of the first volume of the book, Black to play and live. 
In the initial position, Black has only three stones, but the shape and location are favorable, allowing for the creation of two eyes in the narrow area. 
Fig. \ref{fig:prob1_b} shows one move sequence of the solutions.
RZS-TT solved the problem within 184,990 nodes, while in the RZS-PT version, it takes 90,649 nodes, resulting in a speedup of approximately 2.04 times.

This problem is not considered difficult for experienced players. 
However, when using the RZS approach to solve, an unexpected RZ was generated, as shown by the grey region in Fig. \ref{fig:prob1_d}. 
After examining the solution tree, it was discovered that the RZS actually generated the move sequence as shown in Fig. \ref{fig:prob1_c}. 
But in fact, Black did not need to continue atari the White stones after move 15. 
Instead, Black 15 could play the delayed atari move at the location of 17 (the same technique as in 19) and finally capture the White stones, resulting in a smaller RZ. 
Although the minimum RZ is not generated, this problem cannot be solved without the help of RZS, which automatically reduces the search space.

Furthermore, this problem illustrates that when relying on human effort to pre-plan a reduced search space according to the problem design, it is necessary to include a certain amount of space in the upper right. 
If the reduced search space is too large, it will decrease the solving efficiency. 
On the other hand, if the reduced search space is too small, it may deviate from the original problem design and even can not be solved. 
Therefore, determining the appropriate range to be included requires careful consideration and effort.

\subsection{Problem 2}
\label{subsec:problem2}

\begin{figure}[t]
\centering
\subfloat[]{\includegraphics[width=0.33\columnwidth]{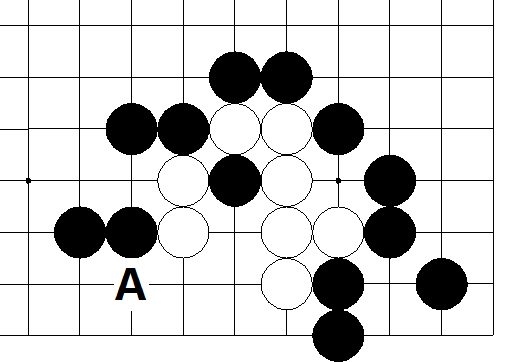}\label{fig:prob2_a}}
\subfloat[]{\includegraphics[width=0.33\columnwidth]{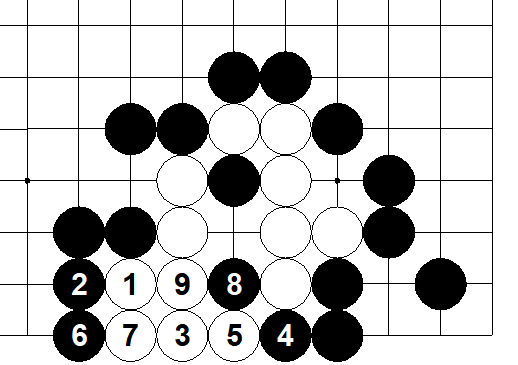}\label{fig:prob2_b}}
\\
\subfloat[]{\includegraphics[width=0.33\columnwidth]{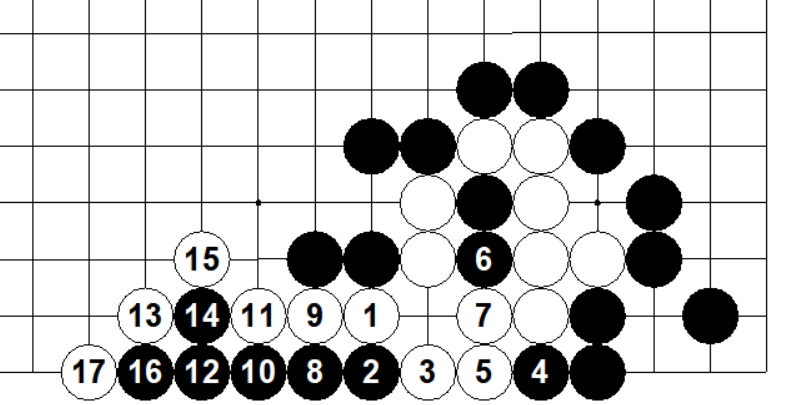}\label{fig:prob2_c}}
\subfloat[]{\includegraphics[width=0.33\columnwidth]{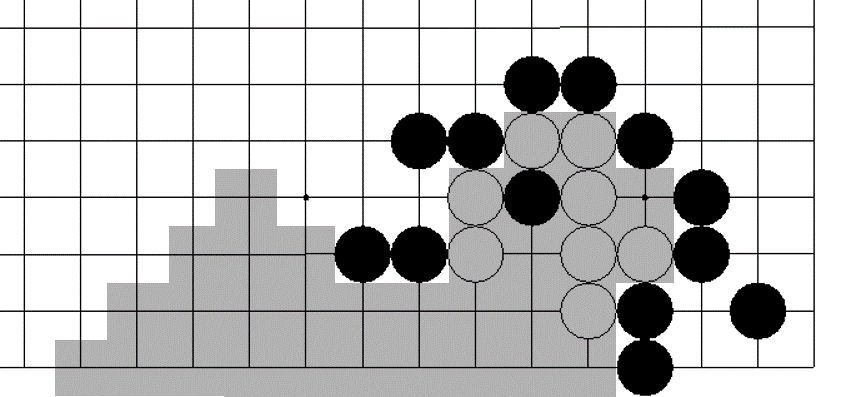}\label{fig:prob2_d}}
\caption{
Problem 2, from Volume 2, page 351 of the book.
}
\label{fig:problem2}
\end{figure}

This problem shown in Fig. \ref{fig:prob2_a} is from page 351 of the second volume of the book, White to play and live. 
White can only expand the area to the left in order to secure a living shape. 
White can only expand the area to the left. Fig. \ref{fig:prob2_b} illustrates the most crucial solution, which leads to a pattern known as \textit{oshitsubushi}, where White lives due to the opponent's illegal move.
In the RZS-TT version, a total of 219,067 nodes were searched, while in the RZS-PT version, 94,313 nodes were searched, resulting in a speedup of approximately 2.32 times.

Surprisingly, in this problem, an elongated RZ was again generated as in Fig. \ref{fig:prob2_d}, even including areas that were not shown in the original book. 
The move sequence that derives this RZ is shown in Fig \ref{fig:prob2_c}. 
This once again demonstrates the advantage of the RZS approach in automatically creating RZ to prune the search, eliminating the need for manual planning and design of the reduced search space.

\subsection{Problem 3}
\label{subsec:problem3}

\begin{figure}[t]
\centering
\subfloat[]{\includegraphics[width=0.33\columnwidth]{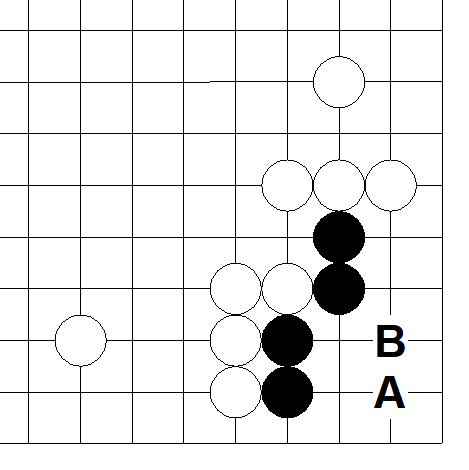}\label{fig:prob3_a}}
\subfloat[]{\includegraphics[width=0.33\columnwidth]{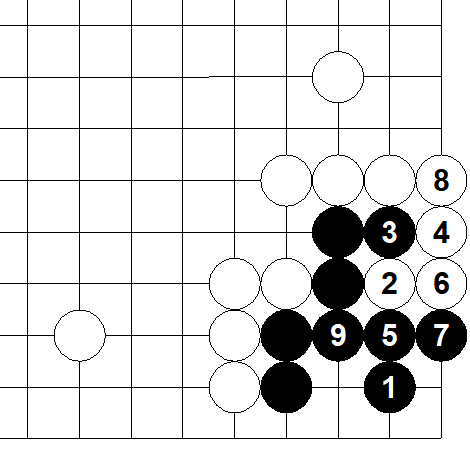}\label{fig:prob3_b}}
\subfloat[]{\includegraphics[width=0.33\columnwidth]{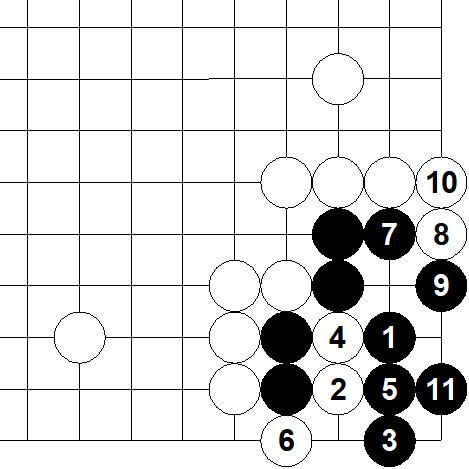}\label{fig:prob3_c}}
\\
\subfloat[]{\includegraphics[width=0.33\columnwidth]{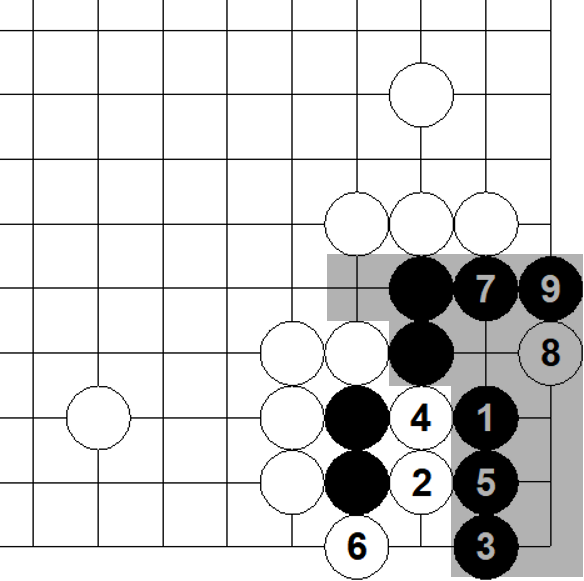}\label{fig:prob3_d}}
\subfloat[]{\includegraphics[width=0.33\columnwidth]{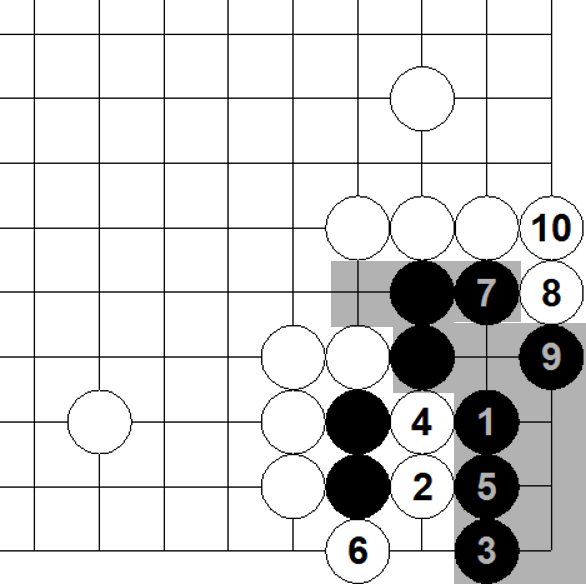}\label{fig:prob3_e}}
\subfloat[]{\includegraphics[width=0.33\columnwidth]{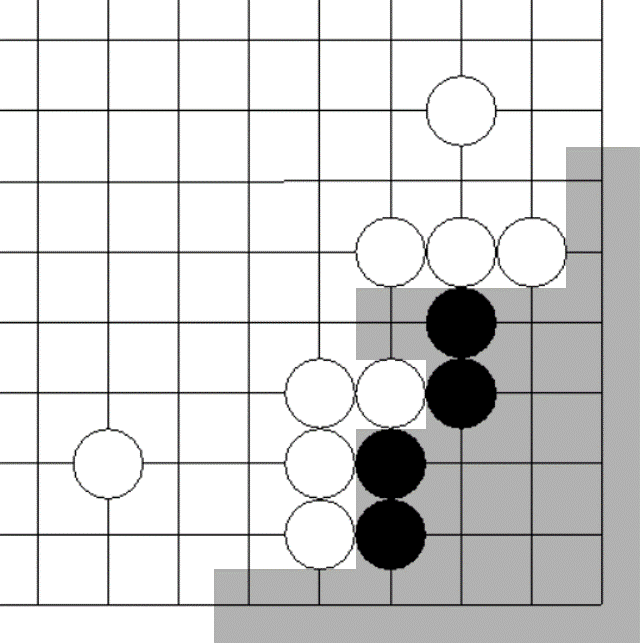}\label{fig:prob3_f}}
\caption{
Problem 3, from Volume 1, page 222 of the book.
}
\label{fig:problem3}
\end{figure}

This problem shown in Fig. \ref{fig:prob3_a} is from page 222 of the first volume, Black to play and live. 
Although Black has a wider area in this problem, their shape is relatively weak, limiting the options for making a living shape. 
RZS-TT solved the problem by A within 450,401 nodes, while the RZS-PT solved it by B and takes 50,212 nodes, a speedup of approximately 8.97 times.

The solvers provided different answers for the problems, but the book \cite{cho_life_1987} only presents the unique correct answer marked as A, while considering B as a failed attempt due to resulting in a smaller territory. 
Fig. \ref{fig:prob3_b} and \ref{fig:prob3_c} display the move sequences corresponding to answers A and B. 
Move B involves sacrificing two stones and securing a smaller area for survival. 
This difference reflects a discrepancy between the solvers and human Go players. 
Human Go players often aim for not only solving the game but also the largest territory, whereas the solvers focus on achieving UCA of any crucial stones. 

The RZS-TT solver achieved a speedup of approximately 9 times, primarily due to solving at B with a smaller area, resulting in a smaller RZ in successor positions and enabling the reuse of more patterns. 
For example, \ref{fig:prob3_d} and \ref{fig:prob3_e} are two cases that can be found during RZS-PT and shared among different positions. 
This means that we don't need to consider patterns outside the RZ. 
The White moves outside RZ can be regarded as null moves for the two patterns.

\subsection{Problem 4}
\label{subsec:problem4}

\begin{figure}[t]
\centering
\subfloat[]{\includegraphics[width=0.33\columnwidth]{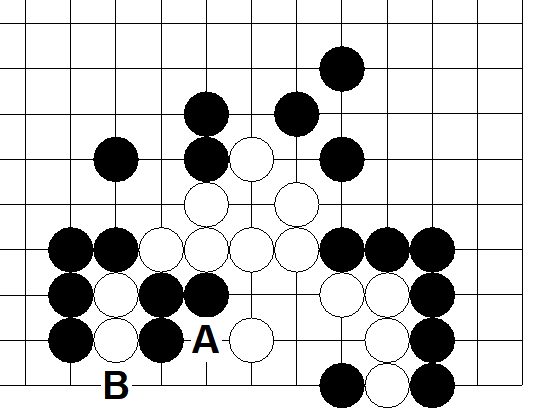}\label{fig:prob4_a}}
\subfloat[]{\includegraphics[width=0.33\columnwidth]{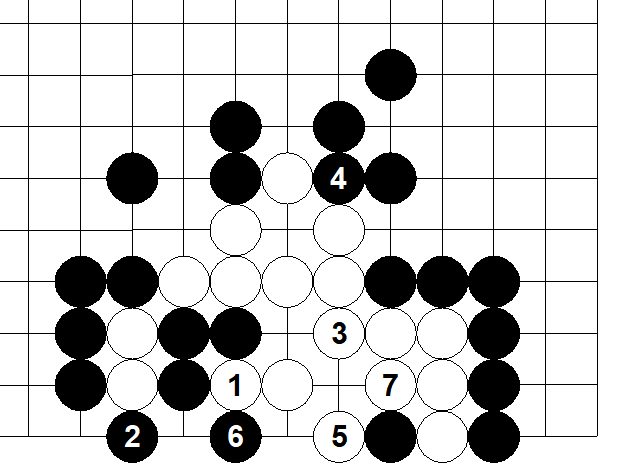}\label{fig:prob4_b}}
\\
\subfloat[]{\includegraphics[width=0.33\columnwidth]{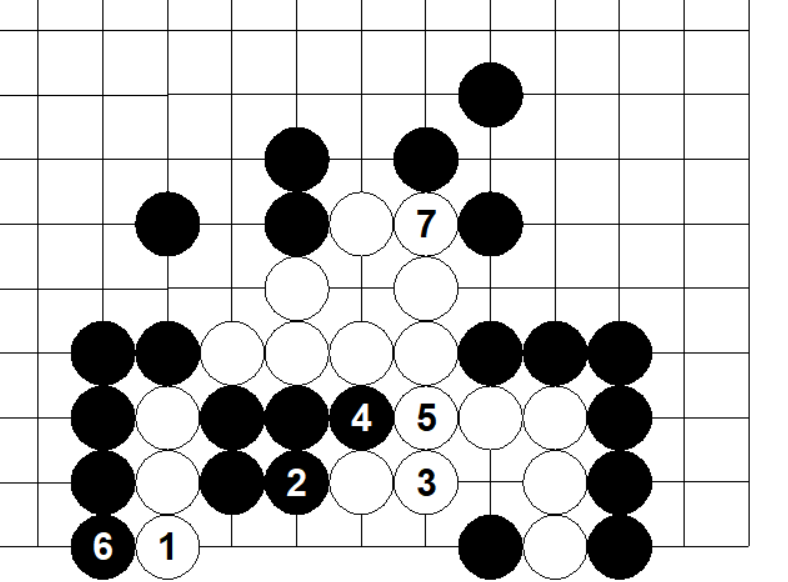}\label{fig:prob4_c}}
\subfloat[]{\includegraphics[width=0.33\columnwidth]{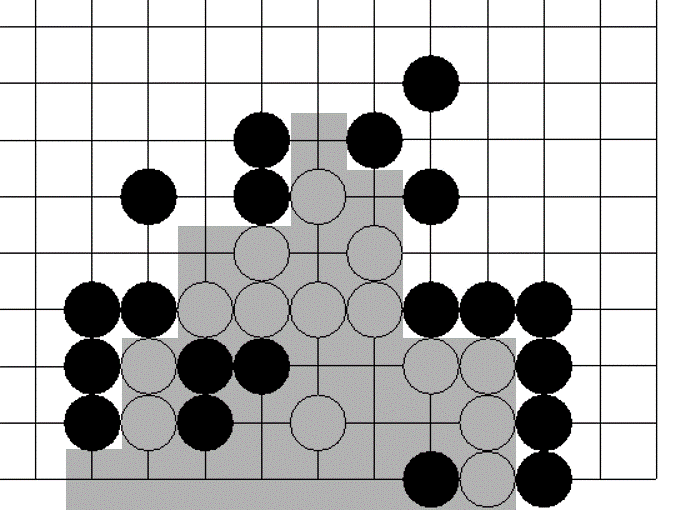}\label{fig:prob4_d}}
\caption{
Problem 4, from Volume 2, page 156 of the book.
}
\label{fig:problem4}
\end{figure}

This problem shown in Fig. \ref{fig:prob4_a} is from page 156 of the second volume of the book, White to play and live. Depending on the situation, White may need to sacrifice the four stones on the right side to live.
RZS-TT solved the problem within 59,898 nodes, while the RZS-PT takes 26,871 nodes, resulting in a speedup of approximately 2.22 times.

Interestingly, the book mentions that the first move for White is difficult and gives the unique correct answer as playing at A. 
Fig. \ref{fig:prob4_b} shows a move sequence of the solutions following move A. 
However, the solvers have discovered that playing at B is also a valid answer. 
Fig. \ref{fig:prob4_c} demonstrates a move sequence following move B. 
It first extends the liberties of the two stones on the left side and then launches an attack on the black stones from the inside. 
It is interesting to see that the solvers have discovered a different and meaningful answer that introduces new thinking and patterns not documented in the book.
This finding also helps Go players explore less intuitive variations that human players may overlook, and provides insights into the differences between human and AI decision-making, offering valuable references for Go pedagogy and improving Go AI training.

It is worth noting that the FTL network provided inaccurate value\footnote{The value estimation of an AlphaZero network on the game of Go is used to estimate how good the given position is in the aspect of the given player to play. 
The values range from -1 to 1, where values closer to 1 indicate a better position, while values closer to -1 indicate a worse position.} estimations for move 1 and move 2 in these two solutions. 
However, after move 3, the network outputs values that align with the desired outcome of White’s winning.

\subsection{Problem 5}
\label{subsec:problem5}

\begin{figure}[t]
\centering
\subfloat[]{\includegraphics[height=2cm]{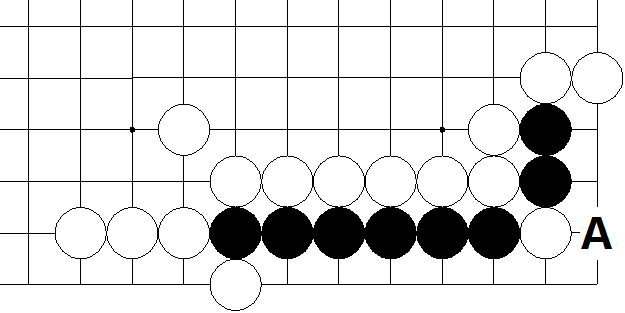}\label{fig:prob5_a}}
\hspace{0.4cm}
\subfloat[]{\includegraphics[height=2cm]{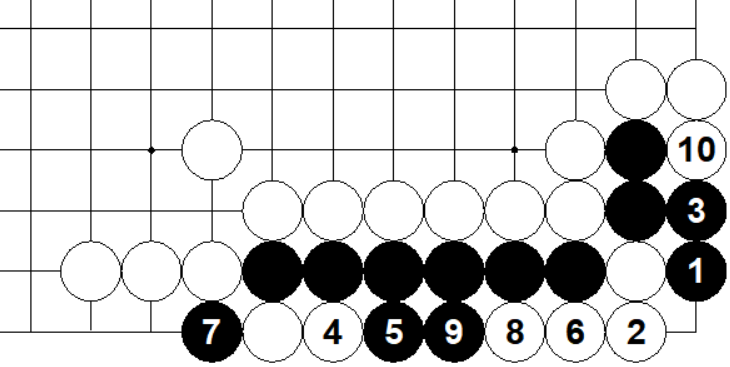}\label{fig:prob5_b}}
\\
\subfloat[]{\includegraphics[height=2cm]{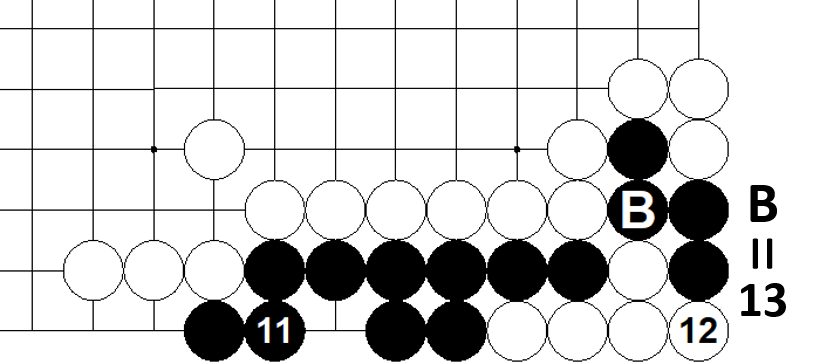}\label{fig:prob5_c}}
\subfloat[]{\includegraphics[height=2cm]{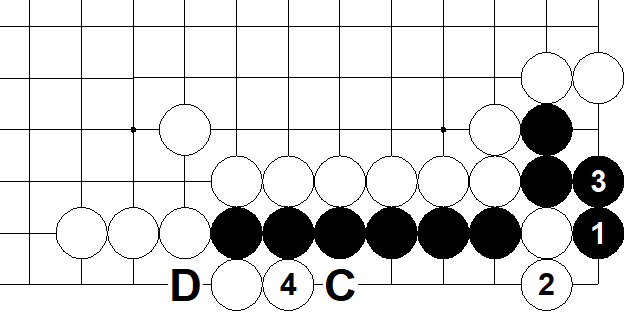}\label{fig:prob5_d}}
\\
\subfloat[]{\includegraphics[height=2cm]{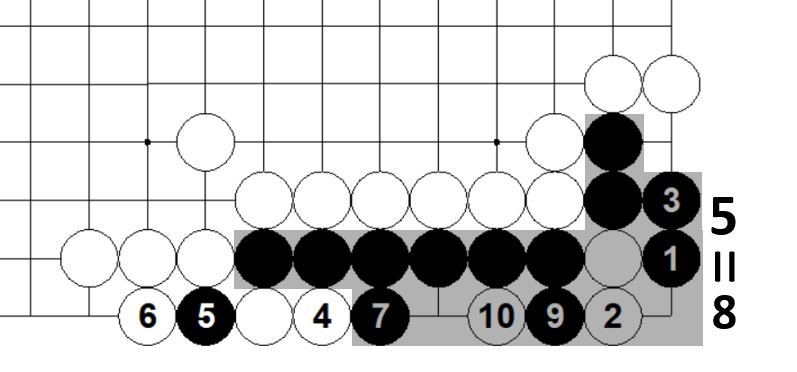}\label{fig:prob5_e}}
\subfloat[]{\includegraphics[height=2cm]{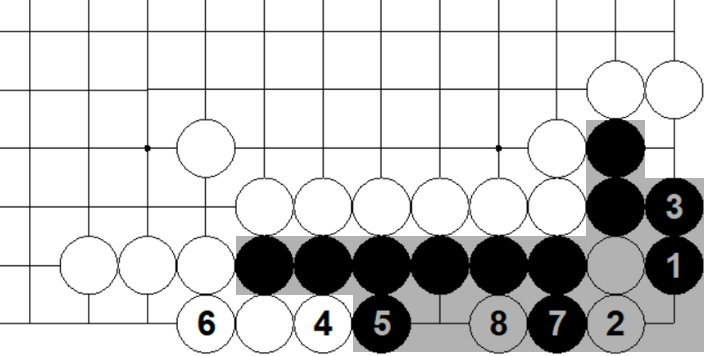}\label{fig:prob5_f}}
\\
\subfloat[]{\includegraphics[height=2cm]{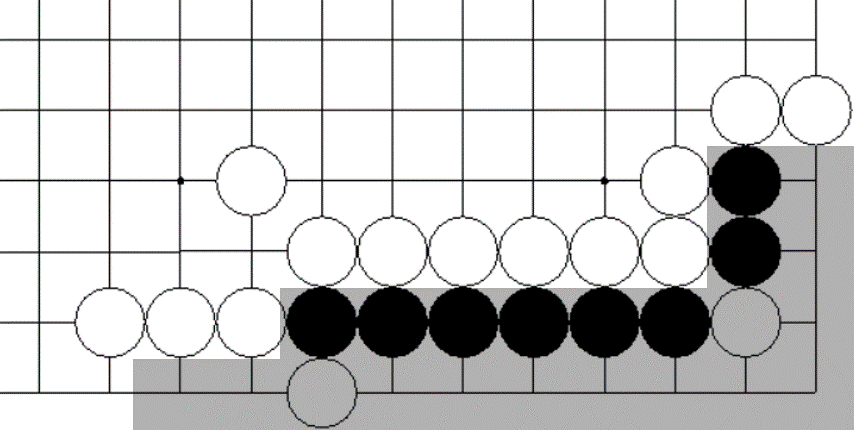}\label{fig:prob5_g}}

\caption{
Problem 5, from Volume 2, page 162 of the book.
}
\label{fig:problem5}
\end{figure}

This problem shown in Fig. \ref{fig:prob5_a} is from page 162 of the second volume of the book, Black to play and live. The problem involves a lightning-style (or z-shape) \textit{ishinoshita} pattern in the corner, with variations that include different numbers of counter-captures. Fig. \ref{fig:prob5_b} and \ref{fig:prob5_c} show the most complex situation for counter-capturing five white stones.
The RZS-TT version required a search of 373,086 nodes to prove the solution, while the RZS-PT version only needed 12,566 nodes, achieving a speedup of approximately 29.69 times.

Although the area of this problem is relatively small, RZS-TT focuses on searching move D (shown in Fig. \ref{fig:prob5_d}) at the fifth move. 
It may initially seem unreasonable, and it requires approximately 300,000 simulations. 
However, playing this move with White's exchange for capturing is a valid strategy, albeit not the optimal one, as it leads to a longer solution for Black's winning outcome. 
The probabilities\footnote{The probability is output from the network's policy head, guiding MCTS on which move to search first. 
From a human perspective, larger probabilities can be seen as our intuitive prioritization of certain locations.} assigned to locations C and D are 0.13 and 0.24, respectively, indicating that MCTS initially prioritized D but ultimately solved the problem with move C. 
This serves as a counterexample to the FTL network.

For RZS-TT, the simulations conducted to search for move D may seem somewhat wasteful. However, for RZS-PT, despite initially searching for move D, the patterns discovered during this search can be reused and ultimately contribute to solving the problem with move C. 
An example of this is illustrated in Fig. \ref{fig:prob5_e}, which represents a pattern found during the search for move D and can be applied to the position encountered during the search for move C as shown in \ref{fig:prob5_f}. 
This reuse of patterns does not occur in RZS-TT as it always matches the entire board. 
This is also why RZS-PT benefits significantly and mitigates incorrect choices in this problem.

\subsection{Problem 6}
\label{subsec:problem6}

\begin{figure}[t]
\centering
\subfloat[]{\includegraphics[width=0.33\columnwidth]{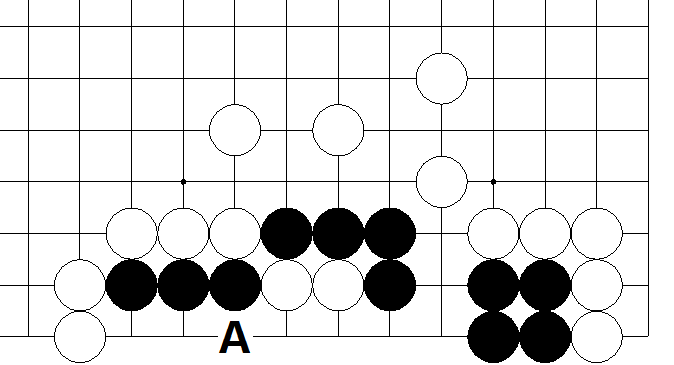}\label{fig:prob6_a}}
\subfloat[]{\includegraphics[width=0.33\columnwidth]{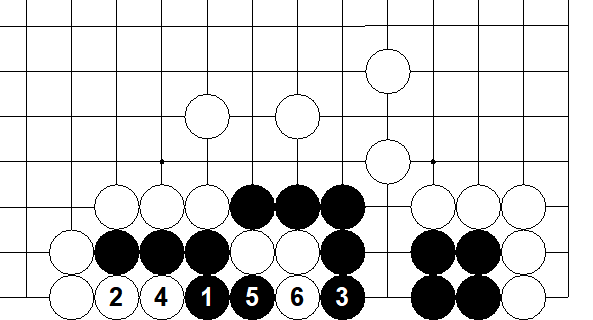}\label{fig:prob6_b}}
\subfloat[]{\includegraphics[width=0.33\columnwidth]{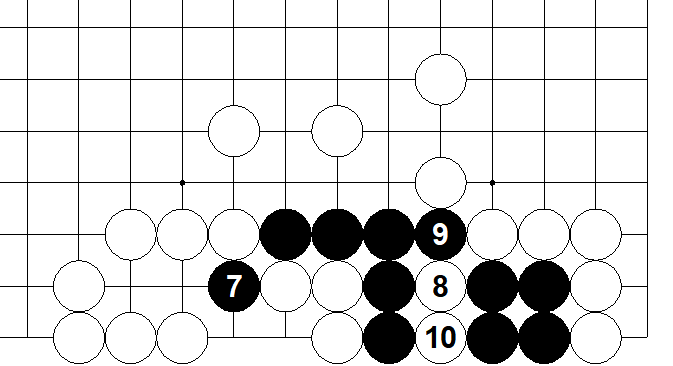}\label{fig:prob6_c}}
\\
\subfloat[]{\includegraphics[width=0.33\columnwidth]{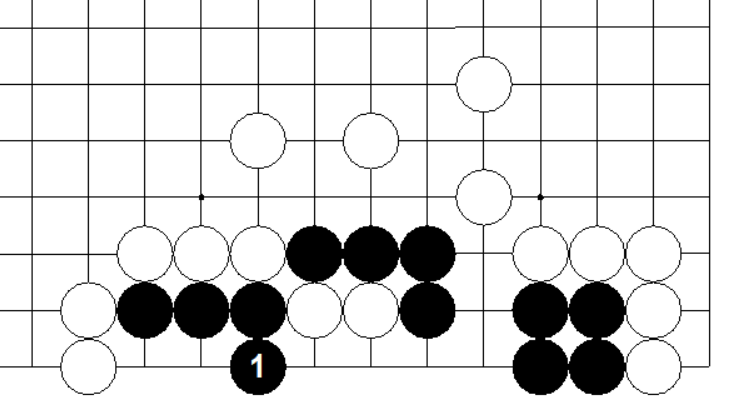}\label{fig:prob6_d}}
\subfloat[]{\includegraphics[width=0.33\columnwidth]{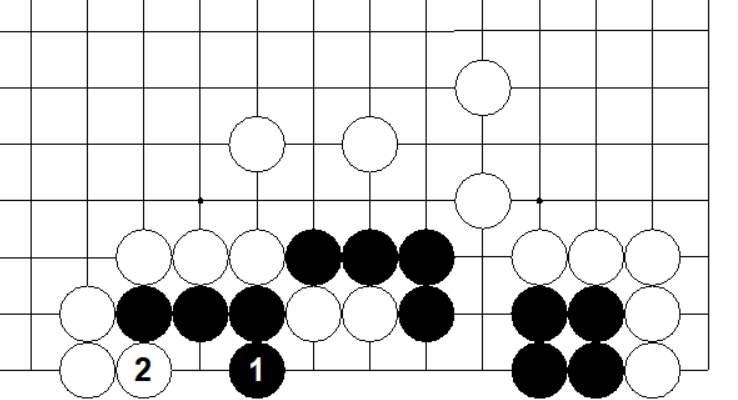}\label{fig:prob6_e}}
\subfloat[]{\includegraphics[width=0.33\columnwidth]{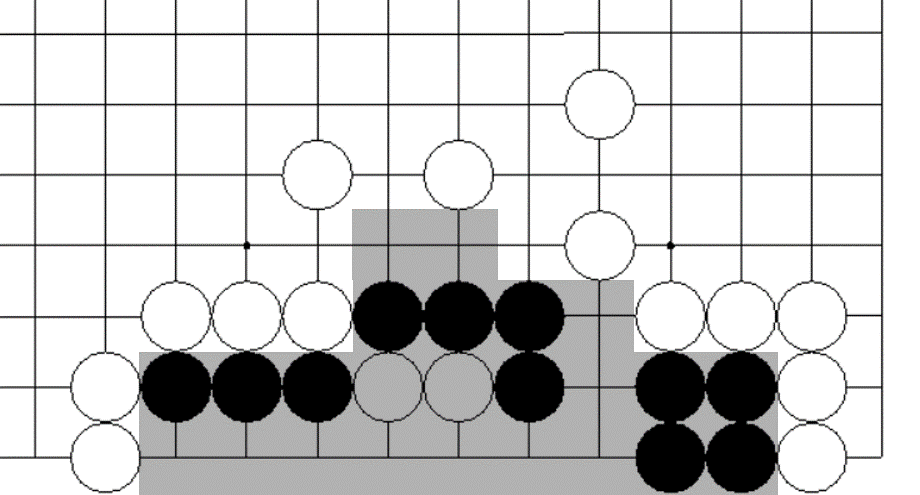}\label{fig:prob6_f}}
\caption{
Problem 6, from Volume 2, page 142 of the book.
}
\label{fig:problem6}
\end{figure}

This problem shown in Fig. \ref{fig:prob6_a} is from page 142 of the second volume of the book, Black to play and live. 
The problem is difficult even for experienced Go players since it contains two rare patterns called \textit{ishinoshita}. 
The \textit{ishinoshita} pattern typically involves allowing the opponent to capture several of one's own stones and then playing a move for capturing the opponent's stones in return. 
Fig. \ref{fig:prob6_b} and Fig. \ref{fig:prob6_c} show the move sequences of the two \emph{ishinoshita} patterns.
RZS-TT solved the problem within 522,342 nodes, while the RZS-PT takes 46,557 nodes, resulting in a speedup of approximately 11.21 times.

It seems to be a difficult problem for the solvers because some positions' value evaluations are completely opposite from the aspect of the player to play. 
For instance, after playing the move shown in Fig. \ref{fig:prob6_d}, the FTL network output the value of 0.99 in White’s aspect. 
In the subsequent position \ref{fig:prob6_e}, the FTL network output a value of -0.61 in Black’s aspect. 
These two cases contradict the expected outcome of Black’s winning. 

Despite initial misjudgments in the early stages, the RZS can still discover the correct moves to succeed by gradually adjusting the Q values\footnote{The Q values play a crucial role in guiding the MCTS to search for promising nodes. 
Ranging from -1 to 1, these Q values are updated by averaging the value estimations from descendant nodes. 
They can also be understood as win rates, giving human Go players an indication of how AlphaZero's MCTS algorithm evaluates a position after some simulations.} through MCTS, and ultimately solve the problem. 
With the help of the RZ pattern table, multiple patterns can be shared among different positions, thereby accelerating the adjustment of Q values and resulting in a speedup of approximately 11 times.

The reason why the network has inaccurate value estimations might be that certain patterns (e.g., ishinoshita) are not sufficiently common in the training data.

\subsection{Problem 7}
\label{subsec:problem7}

\begin{figure}[t]
\centering
\subfloat[]{\includegraphics[height=2cm]{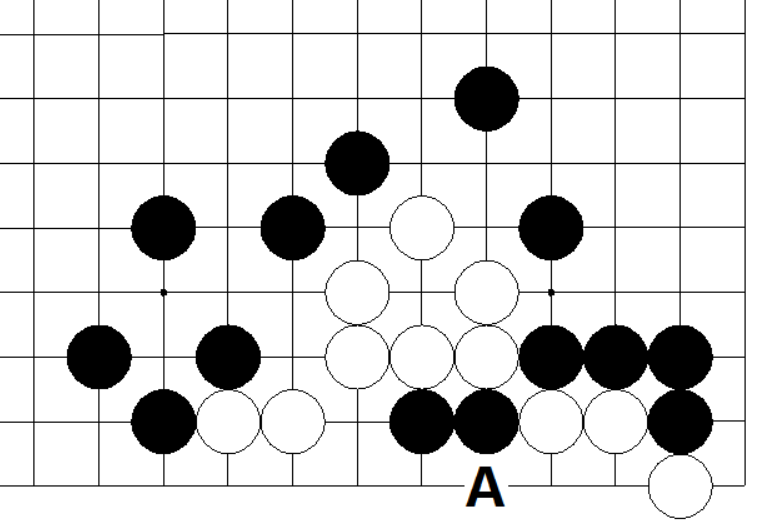}\label{fig:prob7_a}}
\subfloat[]{\includegraphics[height=2cm]{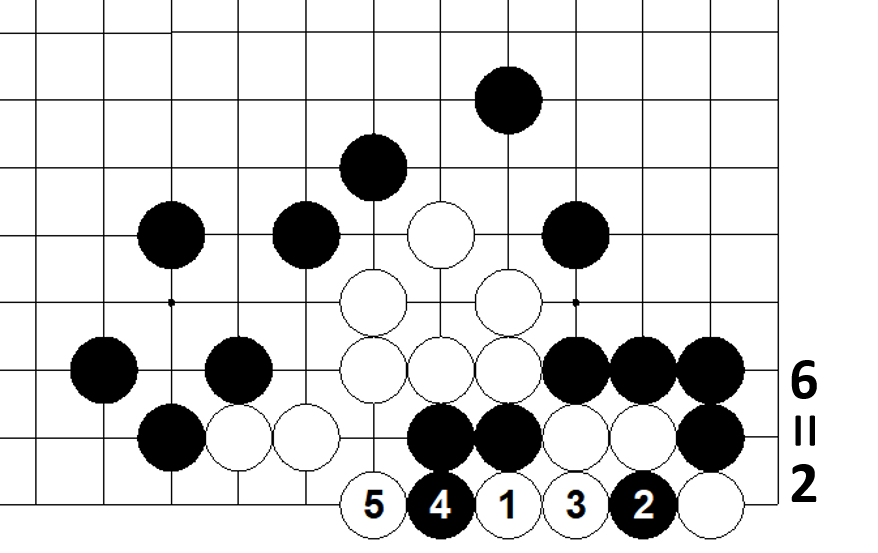}\label{fig:prob7_b}}
\\
\subfloat[]{\includegraphics[height=2cm]{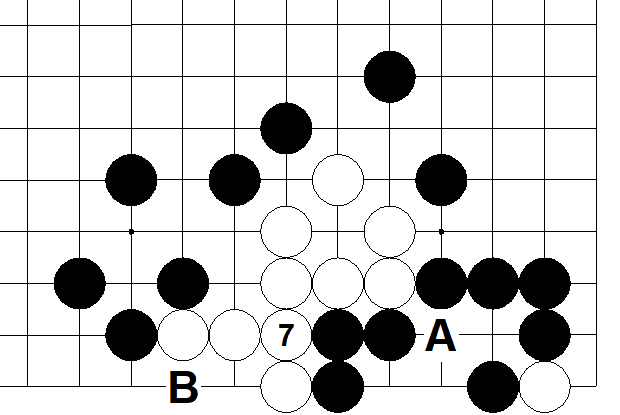}\label{fig:prob7_c}}
\subfloat[]{\includegraphics[height=2cm]{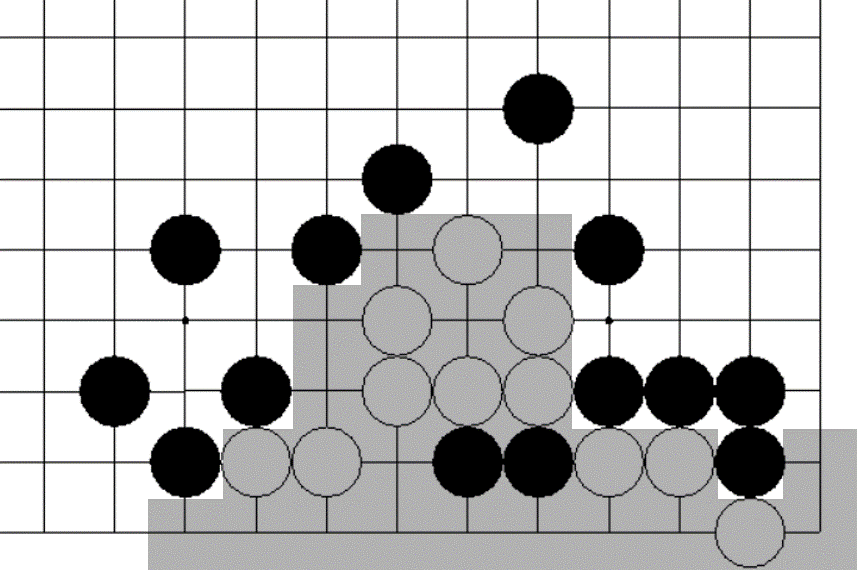}\label{fig:prob7_d}}
\caption{
Problem 7, from Volume 2, page 147 of the book.
}
\label{fig:problem7}
\end{figure}

This problem shown in Fig. \ref{fig:prob7_a} is from page 147 of the second volume of the book, White to play and live. 
It requires sacrificing three stones in the corner to form a pattern similar to ishinoshita pattern. 
The first move is beyond intuition. 
This problem is also recognized as a classic problem from \textit{Xuan-xuan Qi-jing}.

RZS-TT solved the problem within 251,137 nodes, while the RZS-PT takes 22,992 nodes, resulting in a speedup of approximately 10.92 times.

The variant of the move sequence to the \textit{ishinoshita} pattern is shown in Fig. \ref{fig:prob7_b} and Fig. \ref{fig:prob7_c}, White plays 1 in the diagram, then Black counters with 2, White captures with 3, and Black continues with 4, 5, and 6 to capture four white stones. 
After White 7 in \ref{fig:prob7_c}, the A and B points form a miai for both players. 
That is, if Black plays at A, White will play at B to live, and vice versa. 
We also examined the network's probability of playing at A, which ranked second (0.11) among all locations and indicated the network's interest in this move. 
However, the value of the position after move 1 by the network is also misjudged with a value of 0.99 in Black’s aspect. 
Similar to Problem 6, the Q values is finally adjusted to -0.95 in Black’s aspect, which meet the winning outcome of White.

\section{Discussion and Conclusion}
\label{sec:discussion}

In this paper, we conducted a review of the main algorithms employed in \cite{shih_novel_2022,shih_localpattern_2023}. 
Additionally, we presented seven L\&D problems that were successfully solved by the solvers \cite{shih_novel_2022,shih_localpattern_2023}. 
Throughout our analysis, we demonstrated the effectiveness of the RZS in automatically reducing the search space, eliminating the need for manual design of specific areas. 
Furthermore, we identified certain issues related to these problems.
For example, to address the problem of misjudging rare patterns in the training data as mentioned in Problem 6, one potential approach is to perform data augmentation specifically for these problems that involve rare patterns. 
Regarding the issue of not achieving the maximum territories when solving L\&D problems as mentioned in Problem 3, one possible solution is to incorporate additional feature planes that indicate which crucial stones must survive or specify the minimum number of territories the player to live should possess. 
The end game condition should also be modified accordingly.
Other improvements, such as gathering more L\&D problems in the training, remain as future work for further exploration and enhancement.

In addition, while this study focuses on analyzing the behavior of the Go solver in solving L\&D problems, the proposed RZ is general and can be extended to other board games.
Similar analyses, including examining problems where the relevance zone extends far from the initial configuration, cases with alternative correct answers, or problems involving rare patterns, could also be applied to games such as Hex, Slither, Gomoku, and Connect6, providing deeper insights into planning methods in game AI.

\bibliography{references}
\bibliographystyle{IEEEtran}

\end{document}